Running Head:  PROBABILISTIC ANALYSIS OF LANGUAGE ACQUISITION

The probabilistic analysis of language acquisition:

Theoretical, computational, and experimental analysis

Anne S. Hsu

Department of Cognitive, Perceptual and Brain Sciences

University College London

26 Bedford Way

London, WC1H 0AP

ahsu@gatsby.ucl.ac.uk

Nick Chater

Department of Cognitive, Perceptual and Brain Sciences

and ESRC Centre for Economic Learning and Social Evolution (ELSE)

University College London

26 Bedford Way

London, WC1H 0AP

n.chater@ucl.ac.uk

Paul Vitányi

Centrum Wiskunde & Informatica ,

Science Park 123, 1098 XG Amsterdam,

The Netherlands

paul.vitanyi@cwi.nl

**Abstract**

There is much debate over the degree to which language learning is governed by innate language-specific biases, or acquired through cognition-general principles. Here we examine the probabilistic language acquisition hypothesis on three levels: We outline a novel theoretical result showing that it is possible to learn the exact *generative model* underlying a wide class of languages, purely from observing samples of the language. We then describe a recently proposed practical framework, which quantifies natural language learnability, allowing specific learnability predictions to be made for the first time. In previous work, this framework was used to make learnability predictions for a wide variety of linguistic constructions, for which learnability has been much debated. Here, we present a new experiment which tests these learnability predictions. We find that our experimental results support the possibility that these linguistic constructions are acquired probabilistically from cognition-general principles.

**Introduction**

Children learn language primarily from experience, rather than instruction. Indeed, many researchers have suggested that children are able to learn language from mere exposure, without relying on other's feedback about their own utterances. That is, children may be able to learn language primarily from positive evidence alone.

The differences between two main perspectives on language acquisition can be understood through the distinction between *discriminative* and *generative* models of learning (Hsu & Griffiths, 2009). From the point of view of discriminative learning models, learning any pattern from positive examples alone seems deeply puzzling. A discriminative model frames the learning problem as finding a mapping from inputs to categories, from a set of input-category pairs. But if all inputs that the learner encounters are members of the *same* category (i.e., the category of grammatical sentences), then there seems no basis on which to determine the boundaries of this category. Discriminative models are widespread in models of cognition, ranging from the Rescorla-Wagner model of classical conditioning, to feed-forward neural networks, to support vector machines (Jäkel, Schölkopf & Wichmann, 2009). Theoretical analyses of the problem, i.e. learning the set of grammatical sentences from input containing only grammatical examples, tend to be discouraging (e.g., Gold, 1967; Niyogi, 2006). Indeed, thinking about language learning as a classification problem has led many theorists to conclude that language acquisition faces fundamental "logical" problems (Baker & McCarthy, 1981; Hornstein & Lightfoot, 1981).

The Bayesian approach to cognitive development, explored in this special issues, and the cognitive sciences more generally (e.g,, Griffiths, Chater, Kemp, Perfors & Tenenbaum, in press), suggest a different perspective: that the primary task of the learner is to make inferences about the generative model that may have produced the language input. Thus, from this generative perspective, language acquisition is not primarily a matter of discriminating "good" from "bad" linguistic forms; instead the aim is to model the underlying regularities that give rise to the language. The key assumption for generative learning is that the input is sampled from the natural language distribution; discriminative learning models do not require this assumption.

In this paper, we present a generative Bayesian perspective on the problem of language acquisition spanning three levels of analyses, theoretical, computational and experimental. The theoretical and experimental results are novel contributions of this paper and the computational results are summarized from recent related work. First, we present new positive theoretical learnability results, indicating that under fairly general conditions, it is possible to precisely identify the generative language model (assuming a sufficiently long typical sample of positive data). Combined with prior work, these results suggest that the "logical" problem of language acquisition may be reduced by adopting a probability perspective. Second, we review a recently proposed, general framework which quantifies learnability of constructions in natural language. This framework, based on a particular instantiation of the Bayesian approach known as the simplicity principle, has been used to predict natural language learnability for a wide variety of linguistic rules, using corpus data. Third, we present a new experiment which tests these learnability predictions by comparing them with adult grammaticality judgments.

## Gold revisited: Generative model identification in the limit

A central theoretical question is: given sufficient exposure to the language, can the learner recover a perfectly accurate description of that language? Gold (1967) famously showed that, under certain assumptions, this is not possible. However, a range of more positive results have since been derived, e.g., (Chater & Vitányi, 2007; Feldman et al., 1969). These results apply across linguistic levels: including the acquisition of phonology, morphology, syntax, or the mapping between syntax and logical form.

Here, we outline a new and strong positive learnability result. Our most basic result is the *Computable Probability Identification Theorem*. Although the result applies more generally, we will frame the discussion in terms of language. Informally, the theorem can be stated as follows: suppose an indefinitely long language corpus is generated by identical independently distributed (i.i.d.) samples from some probability distribution, *p*, over a countable set (e.g., the set of sentences in a language). We require only that *p* is computable[1]: there exists some computational device (e.g., a Turing machine), that, for each *x*, can compute the probability *p(x)*.

Then there exists a computable algorithm that receives the sentences, in sequence, and generates a sequence of guesses concerning the generative probabilistic model of the language (see Appendix). We will call these guesses $q_1, q_2,...$ After sufficient linguistic data, the algorithm will almost surely alight on a guess which subsequently never changes: that is the sequence $q_1, q_2,...$ "converges" almost surely to *q*. Moreover *q=p*, the probability distribution generating the language. This implies that the learner can then itself *generate* language using the correct probability distribution.

This result indicates not only that there need not be a "logical" problem of language acquisition (Baker & McCarthy, 1981; Hornstein & Lightfoot, 1981); but provides an algorithm which defines a computable process that will precisely identify not only the language, but the precise generative probabilistic model from which the language is sampled. This could be viewed as a Universal Induction Algorithm---note, in particular, that this algorithm embodies no knowledge specific to language.

This result is stronger than many previous results, in a number of ways. (1) The language can be learned from the entire class of computable generative probability distributions for language. Thus, the method is not restricted to particular classes of language structure, such as bigram models, hidden-markov models or probabilistic context-free grammars. In contrast, many theoretical results are specific to a particular model class (e.g., Feldman et al., 1969; Manning & Schütze, 1999). (2) The learner does not merely approximate the language, as in most probabilistic results, but identifies the generating distribution precisely. (3) The learning method is computable (unlike, for example, Chater & Vitányi, 2007).

A number of open questions remain. The proof in the i.i.d. case depends on the strong law of large numbers. The question remains whether our results hold under weakened i.i.d. assumptions, e.g. sentences which have only "local" dependencies (relative to the total corpus the learner has been exposed to). In reality, there are complex interdependencies between sentences at many scales. These may arise from local discourse phenomena such as anaphora, to much higher-level dependencies determined by the overall coherence of a conversation or narrative. One possibility is that these dependencies will "wash out" over long time horizons. Additionally, for probabilistic processes which are stationary and ergodic, there are limit theorems analogous to the strong law of large numbers, raising the possibility that analogous results apply. Note that the present result is much stronger than traditional traditional language identification in the limit (e.g., Osherson, Stöb & Weinstein, 1985): we show that the precise probability distribution generating language can be precisely identified, not merely the set of sentences allowed in the language. A second open question concerns the number of sentences typically required for identification to occur. We leave these, and other, open questions for a later technical paper (Vitányi & Chater, in preparation).

The present result might appear remarkably strong. After all, it is not generally possible almost surely to precisely identify the probability $p$ with which a biased coin lands heads from a finite set of coin flips, however long[2]. The same applies to typical statistical language models, such as probabilistic context-free phrase structure grammars. These language models are typically viewed as having real valued parameters, which is a psychologically and computationally unrealistic idealisation that makes the problem of generative model identification unnecessarily difficult. In practice, any computational process (e.g, inside the head of the parent, to whom the child is listening) can only be determined by *computable* processes—and hence computable parameters, dramatically reducing the possible parameter values. We do not mean to suggest that the child can or does precisely reproduce the generative probabilistic model used by adult speakers. But if such identification is possible almost surely, for any computable linguistic structure, the child presumably faces no insurmountable logical problems in acquiring a language from positive data alone.

## A practical framework for quantifying learnability

We have presented above a new and strong learnability result. But much debate concerning language acquisition concerns more specific questions, such as how children learn restrictions to general rules from exposure to only positive language examples. Restrictions on the contraction of '*going to*' provide an illustrative example: '*I'm gonna leave*' is grammatical, whereas '*I'm gonna the store*' is ungrammatical. Language acquisition requires the speaker to generalize from previously heard input. Research indicates that many (perhaps most) children are rarely corrected when they produce an over-general, ungrammatical sentence. Children also are not explicitly told which generalizations are allowed (Bowerman, 1988). These observations evoke the question: how do children learn which overgeneralizations are ungrammatical without explicitly being told? Many language acquisition researchers have traditionally claimed that such learning is impossible without the aid of innate language-specific knowledge (Chomsky, 1975; Crain, 1991; Pinker, 1989).

The learnability results we present above counter such a priori arguments: We show a general solution to the problem acquiring language is possible, under which any specific linguistic structure can be acquired from positive data alone. Moreover, researchers have shown that concrete statistical models are capable of learning restrictions to general rules from positive evidence only (Dowman, 2007; Foraker et al., 2009; Grünwald, 1994; Perfors et al., 2006; Regier & Gahl, 2004). These statistical models are based on a particular instantiation of Bayesian modelling in which languages are chosen based on the principle of simplicity.

Inherent in a simplicity-based Bayesian approach to language acquisition is the trade-off between simpler vs. more complex grammars: Simpler, over-general grammars are easier to learn. However, because they describe language statistics less accurately, they encode language input less efficiently, i.e. longer code lengths are required to represent language. More complex grammars (which enumerate linguistic restrictions) are more difficult to learn, but they better describe language statistics and result in a more efficient encoding of the language, i.e., language can be represented using shorter code lengths. Because complex grammars become worthwhile as linguistic constructions appear more often, simplicity models are able to learn restrictions from positive evidence alone (See Figure 1).

Recently, a *general quantitative framework* has been proposed which can be used to assess the learnability of any given *specific linguistic restriction* in the context of real language, using positive evidence and language statistics alone (Hsu & Chater, 2010). This framework built upon previous simplicity-based modelling approaches (Dowman, 2007; Foraker et al., 2009; Perfors et al., 2006) to develop a method that is generally applicable constructions in natural language. When using this framework to analyze learnability of a linguistic construction, there are two main assumptions: 1) The description of the grammatical rule for the construction to be learned. 2) The choice of corpus which approximates the learner's input. Given these two assumptions, the framework provides a method for quantifying learnability from language statistics. The framework allows for comparison of different learnability results which arise from varying these two main assumptions, thus providing a common forum for quantifying and discussing language learnability.

*The Minimum Description Length hypothesis*

Because this framework is detailed elsewhere (Hsu & Chater, 2010), we provide a brief overview here. Learnability evaluations under simplicity can be instantiated through the principle of minimum description length (MDL). MDL is a computational tool that can quantify the information available to an idealized statistical learner of language as well as of general cognitive domains (Jacob Feldman, 2000). When MDL is applied to language, grammars are represented as a set of rules, such as that of a probabilistic context free grammar (PCFG) (Grünwald, 1994). An information-theoretic cost can then be assigned to encoding the grammar rules as well as to encoding the language under those rules. MDL has close formal relations to Bayesian probabilistic analysis, although we do not focus on this here (see Chater, 1996 for an informal description; Vitányi & Li, 2000 for a detailed analysis).

Hsu & Chater (2010) used two-part MDL. In the context of language acquisition, the first part of MDL specifies probabilistic grammatical rules to define a generative probability distribution over linguistic constructions, which combine to form sentences. Note that these probabilities do not necessarily mirror the true frequencies of sentences in the language, but the probabilities as specified under the current hypothesized grammar. The second part of MDL uses the probabilistic grammar to encode all the sentences that a child has heard so far. MDL selects the grammar that minimizes the *total* code length (measured in bits) of both the grammatical description and the encoded language length[3].

According to information theory and the MDL principle, the most efficient code occurs when each data element is assigned a code of length equal to the smallest integer greater than or equal to -$\log_2(p_n)$ bits, where $p_n$ is the probability of the *n*th element in the data. For our purposes, these elements are different grammar rules. The probabilities of these grammar rules are defined by the grammatical description in the first part of MDL. Because efficient encoding results from knowing the correct probabilities of occurrence, the more accurately the probabilities defined in the grammar match the actual probabilities in language, the more briefly the grammar will encode the sentences in the language.

Under MDL, the learner prefers the grammatical description that provides the shortest two-part code for the data received so far. Savings occur because certain grammatical descriptions result in a more efficient (shorter) encoding of the language data. If there is little language data (i.e., a person has little language exposure), encoding detailed specification of the language in the first part of the code will not yield large enough savings in the second part of the code to be chosen. Instead, a "cheaper," simpler, grammar will be preferred. When there is more language data, investment in a more costly, complicated grammar becomes worthwhile. This characteristic of MDL learning can explain the early overgeneralizations followed by retreat to the correct grammar that has been observed in children's speech (Bowerman, 1988).

*A Practical example:*

While details are described in Hsu & Chater (2010), we provide a brief example of how the framework is used to assess learnability of the restriction on contraction of *going to* (see above). Learnability is assessed through the two parts of MDL: 1) difference between original and new grammar lengths 2) savings under new grammar per construction occurrence. Here the old grammar allows *going to* to contract under all circumstances. The new grammar will enumerate situations where contraction of *going to* is not allowed. The difference between grammars encoding lengths came from defining the specific situations where *going to* can and cannot contract, i.e. [contractable going to] = [going to] [verb] and [not-contractable going to] [going to] [a place]. Concepts within brackets were represented as a single encoded symbol: [going to] represents use of the words *going to* in a sentence and [verb] represents any verb and [a place] represents any destination one may go to. These formally correspond to the use of *to* as part of an infinitive verb, e.g. *I am going to stay* and the use of *to* as a preposition meaning *towards* , e.g. *I am going to school* [4]. The encoding length of these additional definitions can then be quantified in bits through MDL, see Hsu & Chater (2010).

The second part of learnability requires the evaluation of savings under the new grammar. This requires specifying the occurrence probabilities (estimated from a chosen corpus) of each sentence form under original vs. new grammars. Under the original grammar, where *going to* contractions are always allowed, finite code lengths are required to encode whether contraction occurs in all situations. Under the new grammar, *going to* contraction never occurs when *to* is a preposition and thus 0 bits are required to encode contraction. From the encoding costs under the original vs. new grammars, we can calculate the savings accrued per occurrence of *going to* contracted in the infinitive form (which is the only one where contraction is allowed). The number of occurrences needed to learn the construction is obtained by determining the amount of occurrences needed so that savings becomes greater than or equal to the cost in grammar length difference.

## Testing learnability predictions

Hsu & Chater (2010) used the above framework to predict learnability for linguistic rules whose learnability have been commonly debated in the language acquisition field. These rules all involve restrictions on a more general rule concerning the following 17 constructions[5]: contractions of *want to, going to, is, what is and who is*; the optionality of *that* reduction; dative alternation for the verbs *donate, whisper, shout, suggest, create, pour*; transitivity for the verbs, *disappear, vanish, arrive, come, fall* (see Hsu & Chater (2010) for details). There was a large spread in learnability predictions. Some constructions appeared learnable within a few years whereas others required years beyond human life spans. Hsu & Chater (2010) compared predicted MDL learnablity with child grammar judgments from previous experimental work (Ambridge et al., 2008; Theakston, 2004). It was found that child grammar judgments were better correlated with MDL learnability than with frequency counts. However, the comparison with child judgements was limited to a handful of constructions. Here, we wish to test learnability predictions for the full range of constructions analysed in Hsu & Chater (2010). To do so, we propose that learnability should also correlate with *adult* grammaticality judgments: The easier a construction is to learn, the greater the difference should be between judgments of the ungrammatical vs. grammatical uses of the construction. This method of using relative grammar judgments to test linguistic learnability has been previously used in children (Ambridge et al., 2008).

*Model Predictions*

The most appropriate type of corpus for making learnability predictions is that of child-directed speech, e.g. CHILDES database (Mac Whinney, 1995). However, because many constructions don't occur often enough for statistically significance, Hsu & Chater (2010) analysed only four constructions using CHILDES. Therefore, we use model predictions obtained in Hsu & Chater (2010) using the full Corpus of Contemporary American English (COCA), containing 385 million words (90% written, 10% spoken), a reasonable representation of the distributional language information that native English language speakers receive. Learnability results using the British National Corpus were similar to that from COCA (Hsu & Chater, 2010). Figure 2 shows the estimated number years, $N_{years}$, required to learn the 17 constructions (Hsu & Chater, 2010). $N_{years}$ was calculated as $O_{needed}/O_{year}$ where $O_{needed}$ is the number of occurrences needed for learning under MDL and $O_{year}$ is the number of occurrences per year, estimated from COCA. We quantify learnability as $log(1/N_{years})$. This puts $O_{year}$ in the numerator, allowing for direct comparison with the entrenchment hypothesis which compares grammar judgments occurrence frequency (see below). All frequency estimates were conducted through Mark Davies online corpus analysis site (Davies, 2008). The learnability estimates depend on an assumption of the number of total symbols used in a child's original grammar. Here we present results assuming 100,000 total symbols. The relative learnability does not change depending on the assumed number of total symbols. However, the general scale does change, e.g. when assuming 200 total symbols, number of years needed is approximately halved for all constructions. Thus the learnability results of Hsu & Chater (2010) are best interpreted as quantifiers of relative rather than absolute learnability.

*Learnability vs. entrenchment*

To verify that our experimental results cannot trivially be explained by a simpler hypothesis, we compare our experimental results with entrenchment theory (Brooks et al., 1999). According to entrenchment theory, the likelihood of a child over-generalizing a construction is inversely related to the construction's occurrence frequency. There is some relation between learnability and entrenchment predictions because high construction occurrence frequencies do aid learnability. However, learnability differs from mere frequency counts because MDL also takes into account the complexity of the grammatical rule that governs the construction to be learned, as well as the relative occurrence probabilities of the various forms of each construction (e.g. infinitive vs. prepositional form of *going to*). Thus, while learnability takes frequency into account, frequency and learnability will not necessarily be correlated because of the additional factors that are taken into account by learnability. Here, as in Ambridge et al. (2008), we examine whether relative grammar judgments are related to the construction's input occurrence frequency (frequencies estimated from COCA).

*Experimental method*

*Participants* 105 participants were recruited for an online grammar judgment study (age range: 16-75 years, mean: 34 years). Results were included in the analysis only for participants who were native English speakers (97 out of 105 participants). The majority (74%) of participants learned English in the United States. Other countries included the UK (14%), Canada (5%), Australia (4%).

*Procedure* Participants rated the grammaticality of both grammatical and ungrammatical sentences involving the 17 constructions whose learnability were quantified above. These sentences (34 total) are shown in Table 1. Grammar judgments ranged from 1-5: 1) Sounds completely fine (Definitely grammatical) 2) Probably grammatical (Sounds mostly fine) 3) Sounds barely passable (Neutral) 4) Sounds kind of odd (probably ungrammatical) 5) Sounds extremely odd (Definitely ungrammatical).

*Results*

Results show a strong correlation between averaged relative grammaticality vs. log learnability as predicted by MDL, $r=.35$; $p=.0045$ (see Figure 3). Relative grammaticality for a given linguistic construction is the grammatical rating for the ungrammatical sentence subtracted by the rating for the grammatical sentence. Note that 4 is the maximum possible relative grammaticality because the lowest ungrammatical rating is 5 and the highest grammatical rating is 1. In contrast, grammaticality and construction occurrence frequency are not correlated, as would be expected according to entrenchment, $r=-.08$; $p=.77$ (see Figure 4).

## Summary and Conclusions

This work helps evaluate how much of first language is probabilistically acquired from exposure. We show that, despite putative "logical problems of language acquisition," *any* language generated from any computable generative probability distribution (including any grammars proposed in linguistics) can be precisely identified, given a sufficiently large i.i.d. sample. Our Universal Induction Algorithm embodies no language-specific knowledge, and therefore indicates that language acquisition is possible in principle, given sufficiently large amounts of positive data, and sufficient computing power.

However, how practically learnable are the types of linguistic patterns that have been often cited as challenges for learnability? To address this, we described a recently formulated framework which allows probabilistic learnability to be quantified. Together, these analyses contribute to a substantial body of work showing that probabilistic language learning *is theoretically and computationally possible*.

Does such probabilistic learning occur in practice? Here we propose that if language is probabilistically acquired, then this should leave traces in adult grammar judgments. MDL learnability assumes that a grammar is learned in an absolute sense: once a grammar is chosen under MDL, that is the one used and there is no gradation of knowledge. However, here we conjecture that learnability should not only correlate with how much data is required to learn a linguistic rule, but also the degree of confidence in that knowledge. Experimental results showed that predicted learnability correlates well with relative grammar judgments for the 17 constructions analyzed, chosen as controversial cases from the literature. Our experimental results thus support the possibility that many linguistic constructions that have been argued to be innate may instead be acquired by probabilistic learning.

**Acknowledgements**

This work was supported by grant number RES-000-22-3275 (to Hsu and Chater) from the Economics and Social Research Council, and by the ESRC Centre for Economic Learning and Social Evolution (ELSE). The work of P.M.B. Vitányi was supported in part by the BSIK Project BRICKS of the Dutch government and NWO. Vitányi and Chater were both supported by the EU NoE PASCAL 2 (Pattern Analysis, Statistical Modeling, and Computational Learning).

Footnotes

1. This is a mild restriction, which presumably holds for any cognitively plausible model of language production or mental representation of language; and clearly holds for standard probabilistic models of language, such as probabilistic context free grammar, *n*-gram models, or hidden Markov models.
2. To see this, note that the number of real values on the interval [0,1] is uncountable, whereas the number of guesses associated with any infinite sequence of coin flips is countable. Therefore, the probability any of these guesses is correct has measure 0 in the standard uniform measure on the real interval [0,1].
3. The MDL framework can also be expressed as a corresponding Bayesian model with a particular prior (Chater, 1996; MacKay, 2003; Vitányi & Li, 2000). Here, code length of the model (i.e., grammar) and code length of data under the model (i.e., the encoded language) in MDL correspond to prior probabilities and likelihood terms respectively in the Bayesian framework.
4. These formal definitions are not directly used in learnability analyses because it is unlikely that first language learners are acquiring grammatical knowledge at this level.

5. Hsu & Chater (2010) also included analysis of two more linguistic rules concerning the necessary transitivity of the verbs *hit* and *strike*. Though these verbs are traditionally known to be transitive, in colloquial speech they have evolved to have an ambitransitive usage: e.g. *The storm hit. Lightening struck.* In COCA there are 3678 and 1961 intransitive occurrences of *hit* and *strike* respectively. Thus we did not assess rules regarding the intransitivity of these verbs in our experiment.

Reference List

Ambridge, B., Pine, J., Rowland, C., & Young, C. (2008). The effect of verb semantic class and verb frequency (entrenchment) on children's and adults' graded judgements of argument-structure overgeneralization errors. *Cognition, 106,* 87-129.

Bowerman, M. (1988). The 'No Negative Evidence' Problem: How do Children avoid constructing an overly general grammar? In J.Hawkins (Ed.), *Explaining Language Universals* (pp. 73-101). Oxford: Blackwell.

Brooks, P., Tomasello, M., Dodson, K., & Lewis, L. (1999). Young Children's Overgeneralizations with Fixed Transitivity Verbs. *Child Development, 70,* 1325-1337.

Chater, N. (1996). Reconciling simplicity and likelihood principles in perceptual organization. *Psychological Review, 103,* 566-581.

Chater, N. & Vitányi, P. (2007). Ideal learning' of natural language: positive results about learning from positive evidence. *Journal of Mathematical Psychology, 51,* 135-163.

Chomsky, N. (1975). *The Logical Structure of Linguistic Theory*. London: Plenum Press.

Crain, S. (1991). Language Acquisition in the Absence of Experience. *Behavioral and Brain Sciences, 14,* 597-612.

Davies, M. (2008). The Corpus of Contemporary American English (COCA): 385 million words, 1990-present. Corpus of Contemporary American English [On-line]. Available: http://www.americancorpus.org

Dowman, M. (2007). Minimum Description Length as a Solution to the Problem of Generalization in Syntactic Theory. *Machine Learning and Language, (in review).*

Feldman, J. A., Gips, J., Horning, J. J., & Reder, S. (1969). *Grammatical complexity and inference.* (Rep. No. CS 125). Stanford University.

Foraker, S., Regier, T., Khetarpal, N., Perfors, A., & Tenenbaum, J. B. (2009). Indirect Evidence and the Poverty of the Stimulus: The Case of Anaphoric One. *Cognitive Science, 33,* 300.

Gold, E. M. (1967). Language identification in the limit. *Information and control, 16,* 447-474.

Grünwald, P. (1994). A minimum description length approach to grammar inference. In S.Scheler, Wernter, & E. Rilof (Eds.), *Connectionist, Statistical and Symbolic Approaches to Learning for Natural Language.* (pp. 203-216). Berlin: Springer Verlag.

Hsu, A. & Chater, N. (2010). The logical problem of language acquisition: A probabilistic perspective. *Cognitive Science, in press*.

Hsu, A. & Griffiths, T. (2009). Differencial use of implicit negative evidence in generative and discriminative language learning. *Neural Information Processing Systems, 22*.

Mac Whinney, B. (1995). *The CHILDES project: tools for analyzing talk.* Hillsdale, NJ: Lawrence Erlbaum Associates.

MacKay, D. (2003). *Information Theory, Inference, and Learning Algorithms*. Cambridge: Cambridge University Press.

Perfors, A., Regier, T., & Tenenbaum, J. B. (2006). Poverty of the Stimulus? A rational approach. *Proceedings of the Twenty-Eighth Annual Conference of the Cognitive Science Society,* 663-668.

Pinker, S. (1989). *Learnability and Cognition: The acquisition of argument structure*. Cambridge, MA: MIT Press.

Regier, T. & Gahl, S. (2004). Learning the unlearnable: The role of missing evidence. *Cognition, 93,* 147-155.

Theakston, A. (2004). The role of entrenchment in children's and adults' performance on grammaticality judgment tasks. *Cognitive Development, 19,* 15-34.

Vitányi, P. & Li, M. (2000). Minimum Description Length Induction, Bayesianism, and Kolmogorov Complexity. *IEEE Transactions on Information Theory, IT, 46,* 446-464.

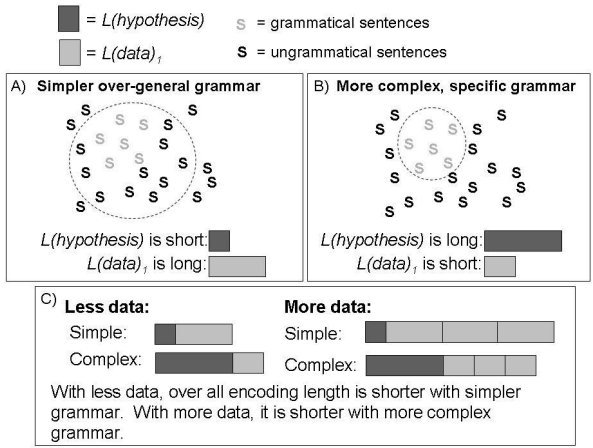

*Figure 1: MDL simple grammar vs. efficient language encoding trade off.* A) A simpler grammar is often over-general, i.e., allows for ungrammatical sentences as well as grammatical ones. Such an over-general grammar may be easy to describe (i.e., short grammar encoding length), but results in less efficient (longer) encoding of the language data. B) A more complex grammar may capture the language more accurately, i.e., allows only for grammatical sentences and doesn't allow for ungrammatical sentences. This more complex grammar may be more difficult to describe (i.e., longer grammar encoding length), but will provide a shorter encoding of language data. C) Initially, with limited language data, the shorter grammar yields a shorter coding length over-all, and is preferred under MDL. However, with more language input data, the savings accumulated from having a more efficient encoding of language data correctly favour the more complex grammar.

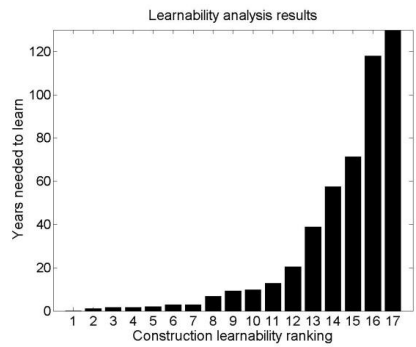

*Figure 2: Estimated years required to learn construction.* Results summarized from (Hsu & Chater, 2010). The constructions are sorted according to learnability: 1) *is* 2) *arrive* 3) *come* 4) *donate* 5) *fall* 6) *disappear* 7) *what is* 8) *shout* 9) *pour* 10) *vanish* 11) *whisper* 12) *create* 13) *who is* 14) *going to* 15) *suggest* 16) *that* 17) **want to*. *Predicted years for learning *want to* is 3,800years.

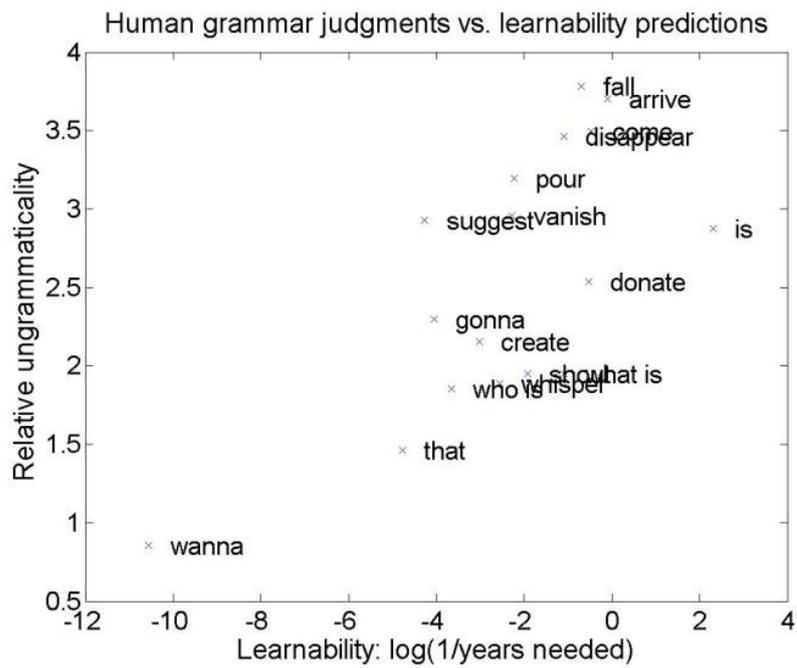

*Figure 3: Human grammar judgments vs. learnability analysis.* Learnability is log of the inverse of the number of estimated years needed to learn the construction. Correlation values: r=.35; p=.0045

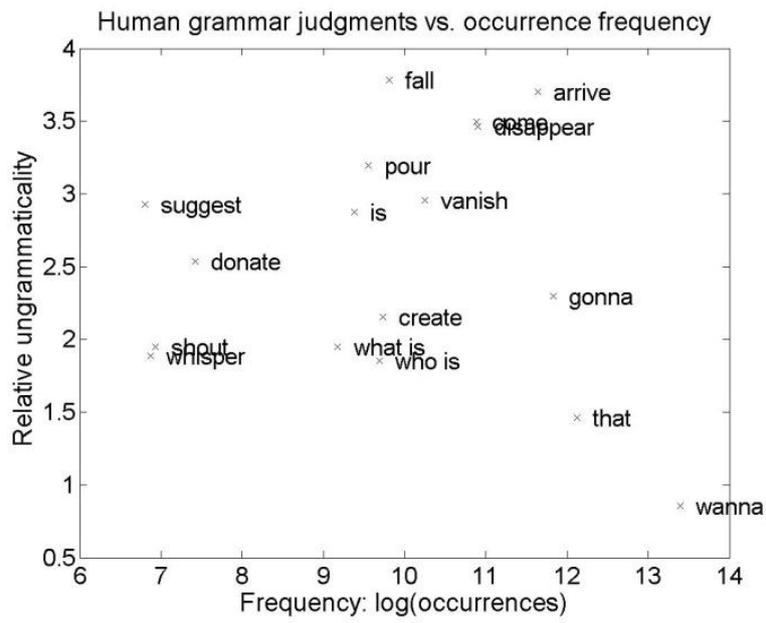

*Figure 4: Human grammar judgments vs. log of occurrence frequency.* Frequencies were estimated using Corpus of Contemporary American English.

*Table 1: Grammatical and ungrammatical sentences used in experiment.*

| Construction | Grammatical usage | Ungrammatical usage |
| --- | --- | --- |
| is | She's as tall as he is. | She is as tall as he's. |
| arrive | The train arrived. | He arrived the train. |
| come | The train came. | I came the train. |
| donate | He donated some money to the charity. | He donated the charity some money. |
| fall | The ornament fell. | He fell the ornament. |
| disappear | The rabbit disappeared. | He disappeared the rabbit. |
| what is | What's it for? | What's it? |
| shout | I shouted the news to her. | I shouted her the news. |
| pour | I poured the pebbles into the tank. | I poured the tank with pebbles. |
| vanish | The rabbit vanished. | He vanished the rabbit. |
| whisper | I whispered the secret to her. | I whispered her the secret. |
| create | I created a sculpture for her. | I created her a sculpture. |
| who is | Who's it for? | Who's it? |
| going to | I'm gonna faint. | I'm gonna the store. |
| suggest | I suggested the idea to her. | I suggested her the idea. |
| that | Who do you think that she called? | Who do you think that called her? |
| want to | Which team do you wanna beat? | Which team do you wanna win? |

# Appendix: Proof of the Computable Probability Identification Theorem

A function is *computable*, if there is a Turing machine (or equivalent) that maps the arguments to the values. Here we consider only probability mass functions wth rational arguments.

The restriction to computable probability mass functions is both cognitively realistic (if we assume language is generated by a computable process) and dramatically simplifies the problem of language identification (for related discussion in a different context, see Cover, 1973).

If a computable function has as values pairs of nonnegative integers, such as $(a, b)$, we can interpret this value as the rational $a/b$. A function $f(x)$ with $x$ rational is *semicomputable from below* if it is defined by a rational-valued computable function $\phi(x, k)$ with $x$ a rational number and $k$ a nonnegative integer such that $\phi(x, k+1) \geq \phi(x, k)$ for every $k$ and $\lim_{k \to \infty} \phi(x, k) = f(x)$. This means that $f$ can be computably approximated arbitrary closely from below (see Li & Vitanyi, 2008, p. 35).

Consider a subclass of the lower semicomputable functions. A function $f$ is a semiprobability mass function if $\sum_x f(x) \leq 1$ and a probability mass function if $\sum_x f(x) = 1$. We write '$p(x)$' for '$f(x)$' if the function is a semiprobability mass function.

It is possible to enumerate all and only the semiprobability mass functions that are lower semicomputable, by fixing an effective enumeration of all Turing machines in a fixed description syntax. Now it is possible to change every Turing machine description in the list into one that computes a semiprobability mass function that is computable from below, as described in the proof of Theorem 4.3.1 in Li & Vitanyi (2008). The list contains all and only semiprobability mass functions that are semicomputable from below.

Every probability mass function is a semiprobability mass function, and every computable probability mass function is semicomputable from below. Therefore, every computable probability mass function is in the list (indeed, each will appear infinitely often).

THEOREM 1 (COMPUTABLE PROBABILITY IDENTIFICATION THEOREM) *Let $L = \{a_1, a_2, \ldots\}$ be a language (a countably finite or infinite set) with a computable probability mass function* $p$. Let the mean of $p$ exist ($\sum_i ip(a_i) < \infty$). Then $p$ can almost surely be computed by an algorithm that takes as input an infinite sequence $x_1, x_2, \ldots$ of elements of $L$ drawn according to $p$.

PROOF. Our data is, by assumption, an i.i.d. sample from a computable probability mass function $p$. Formally, the data $x_1, x_2, \ldots$ is generated by a sequence of random variables $X_1, X_2, \ldots$, each a copy of a single random variable $X$ with probability mass function $P(X = x) = p(x)$ for every $x \in L$. We assume that the mean of $X$ exists (**??**). By the above arguments, we can effectively enumerate the semiprobability mass functions that are computable from below
$$Q = q_1, q_2, \ldots,$$
where $p = q_k$ with least $k$.

DEFINITION 1 *For every $i$ with $1 \leq i < k$, $\max_{x \in L} |p(x) - q_i(x)| > 0$ (this follows from the minimality of $k$). Define*
$$\alpha = \min_{1 \leq i < k} \max_{x \in L} |p(x) - q_i(x)|,$$
*and by definition $\alpha > 0$. For $x^i$ reaches the maximum in $\max_{x \in L} |p(x) - q_i(x)|$, define $t^i$ by $q_i(x^i) - q_i^{t_i}(x^i) \leq \alpha/2$ for $1 \leq i \leq k$. Considering the elements of $L$ lexicographic length-increasing ordered, define $\chi$ by $\chi = \max_{1 \leq i < k} x^i$. Finally, define $\theta$ by $\theta = \max_{1 \leq i \leq k} t^i$.*

We now turn to a probabilistic law that makes it possible to compute index $k$ almost surely given data $x_1, x_2, \ldots$. The Strong Law of Large Numbers states that if we perform the same experiment a large number of times, then almost surely the average of the results goes to the expected value.

Let $\#a(x_1, x_2, \ldots, x_n)$ be the number of elements in $x_1, x_2, \ldots, x_n$ equal $a$ ($a \in L$). Consider some $x \in L$. Then we can consider a Bernoulli process $(q, 1-q)$ where $q = p(x)$ and $1 - q = \sum_{y \in L - \{x\}} p(y)$. Then (Feller, 1968, p 258 ff), for every pair $\epsilon, \delta$, there is an $N$ such that for every $r > 0$ all $r + 1$ inequalities:
$$|p(x) - \frac{\#x(x_1, x_2, \ldots, x_n)}{n}| \leq \epsilon, \tag{1}$$
with $n = N, N+1, \ldots, N+r$ will be satisfied with probability at least $1 - \delta$. That is, we can say, informally, that with overwhelming probability the left hand part of (**??**) remains small for all $n \geq N$. This holds since our sequence of variables $X_1, X_2, \ldots$ satisfies Kolmogorov's criterion that
$$\sum_i \frac{(\sigma_i)^2}{i^2},$$
where $\sigma_i^2$ is the variance of $X_i$ in the sequence of mutually independent random variables $X_1, X_2, \ldots$. Since all $X_i$'s are copies of a single $X$, all $X_i$'s have a common distribution $p$.

We use the theorem on page 260 in Feller (1968). To apply the Strong Law in this case it suffices that the mean of $X$ exists. Thus, if we order the elements of $L$ length-increasing lexicographic, and $i(x)$ is the index of $x$ in the ordered $L$, then we require that

$$\mu = \sum_{x \in L} i(x) p(x) < \infty. \tag{2}$$

Our current "guess" concerning the language can then always be an element in the list of possible lower semicomputable semiprobability mass functions $Q$ not yet ruled out by the data. Since the elements of $Q$ are lower semicomputable, if for some $i$ at some step $t$ we have $q_i^t(x) > p(x)$ then we can rule out $q_i$. But if $q_i^t(x) < p(x)$ it can be the case that at some later step $t' > t$ we have $q_i^{t'}(x) = p(x)$. Thus, our guess of which $q_i$ is actually $p$ may change with the number of steps $t$ and in fact the candidates output may change from earlier on in the list $Q$ to later on to earlier. However, eventually we will identify the correct hypothesis $p$ and this true hypothesis will never be eliminated, however much data is obtained. Thus the true probability distribution is identified. Moreover, it can be shown that this process can be carried out by a concrete algorithm (Vitanyi Chater, in preparation).

●